\documentclass[10pt,twocolumn,letterpaper]{article}

\usepackage{cvpr}
\usepackage{times}
\usepackage{epsfig}
\usepackage{graphicx}
\usepackage{amsmath}
\usepackage{amssymb}
\usepackage{textgreek}
\usepackage{caption}
\usepackage{subcaption}
\usepackage[accsupp]{axessibility}  

\providecommand{\keywords}[1]
{
  \small	
  \textbf{\textit{Keywords---}} #1
}
\PassOptionsToPackage{hyphens}{url}

\usepackage[pagebackref=true,breaklinks=true,colorlinks,bookmarks=false]{hyperref}

\cvprfinalcopy 

\ifcvprfinal\fi

\begin{document}

\title{DIREG3D: DIrectly REGress 3D Hands from Multiple Cameras} 

\author{Ashar Ali\\
\and
Upal Mahbub\\

\and
Gokce Dane\\

\and
Gerhard Reitmayr\\


\and
Qualcomm Technologies, Inc., San Diego, CA, USA.\\
{\tt\small asharali, umahbub, gokced, gerhardr@qti.qualcomm.com}\\

}
\maketitle

\begin{abstract}
In this paper, we present \textit{DIREG3D}, a holistic framework for 3D Hand Tracking. The proposed framework is capable of utilizing camera intrinsic parameters, 3D geometry, intermediate 2D cues, and visual information to regress parameters for accurately representing a Hand Mesh model. Our experiments show that information like the size of the 2D hand, its distance from the optical center, and radial distortion is useful for deriving highly reliable 3D poses in camera space from just monocular information. Furthermore, we extend these results to a multi-view camera setup by fusing features from different viewpoints.
\end{abstract}



\section{Introduction}
Real-time 3D hand tracking from cameras has significant applications in the fields of augmented, virtual and mixed reality \cite{Masurovsky2020ControllerFreeHT}, human-machine interaction, advanced features in automotive and healthcare industry, consumer electronics and vision-based tele-operation of robots. Published solutions for 3D hand modeling use many different types and number of sensors such as IR based sensors, depth sensors \cite{taylor2016efficient}, monochrome or RGB, monocular or multi-view, and multi-modal sensor systems. Yet, in order to build a reliable end-to-end system, major challenges are far from being solved. For example, some solutions require explicit overlap of two or more cameras' fields of view (FOVs) \cite{MegaTrack_FB_2020}. In other solution the power consumption of the chosen sensor modality \cite{issc18} is too large for low power solutions.

\begin{figure*}
    \centering 
    \begin{subfigure}{0.4\textwidth}
    \centering 
    \includegraphics[width=0.9\linewidth]{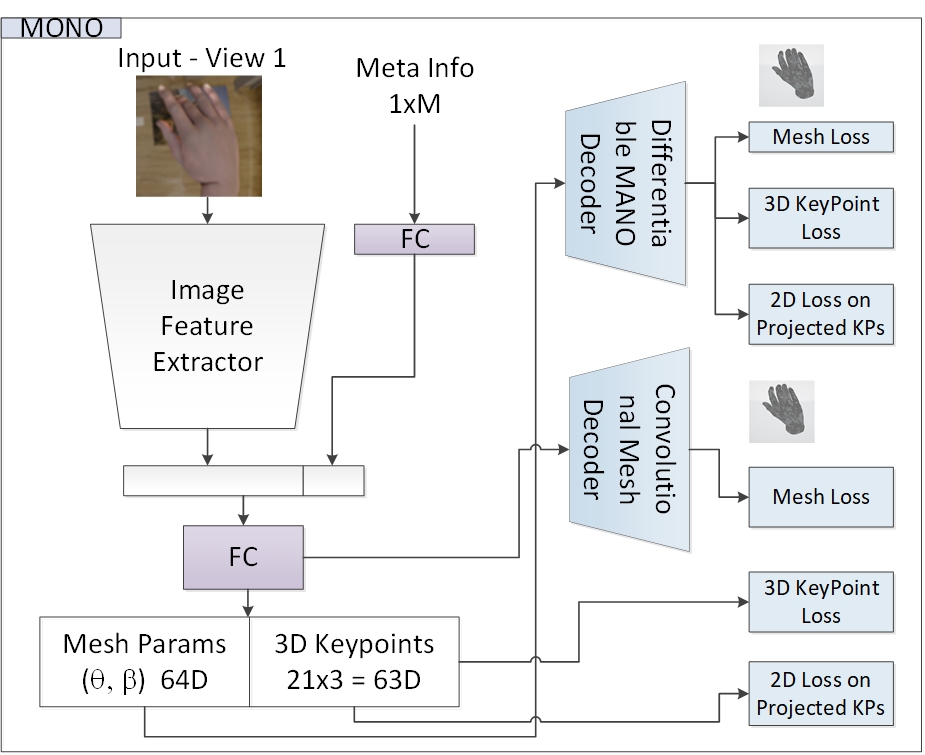}
    \caption{Monocular flow}
    \label{fig:monocular_flow}
    \end{subfigure}%
    \begin{subfigure}{0.6\textwidth}
    \centering 
    \includegraphics[width=0.9\linewidth]{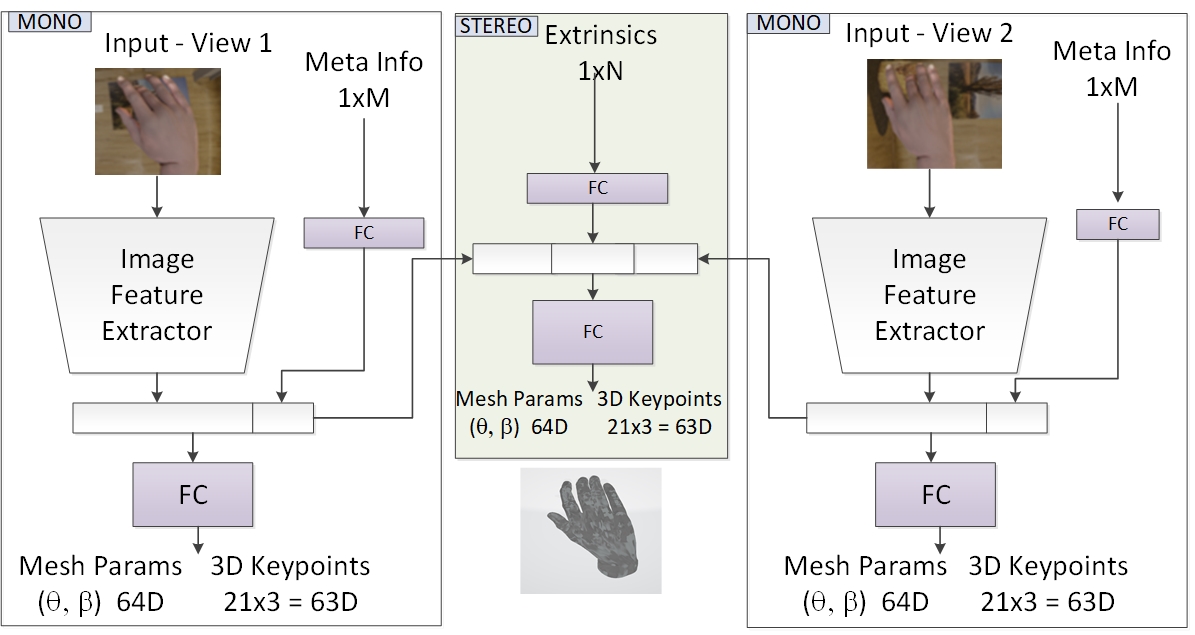}
    \caption{Stereo inference}
    \label{fig:pipeline_flow}
    \end{subfigure}
    \caption{(a) Monocular Case. While training, we jointly optimize for meshes, 3D and derived 2D keypoints.  For inference, only the CNN forward pass provides hand parameters and keypoints directly. (b) Inference in monocular and stereo cases. }
    \vspace{-3mm}
    \label{fig:flow}
\end{figure*}

Our framework addresses these issues with a novel flexible design to achieve the best 3D  modeling performance for both single and multiple view systems with a low-power on-device implementation. Following are the three major contributions of the proposed approach: 
\begin{itemize}
    \item The proposed method is a lightweight framework for directly regressing 3D keypoints, hand mesh, and/or parametric representation of a hand from an input image and a set of meta-information. 
    \item The method incorporates high-level feature fusion to take advantage of multiple views of the same hand in a multi-camera system. 
    \item The novel architecture of the proposed system also enables effectively combining the advantages of parametric and non-parametric estimations using a unique loss function.
\end{itemize}
Hence, we propose a strong and efficient method to directly regress 3D coordinates and mesh parameters from 2D visual cues. Figure~\ref{fig:flow} shows the inference pipeline in more detail. The proposed system can also take crucial meta-information from each camera as input such as the 2D size and location of a hand in a frame, radial distortion of the lens, and 3D rotation between the original optical axis of the lens and center of the bounding box. We also demonstrate the performance gain for expanding into multi-view systems by high-level feature fusion from different views. In addition, we introduce an interesting mechanism to combine the effectiveness of parametric and non-parametric models, along with a unique loss function setup that robustly prepares us for the different variations in data domains, like ground-truth mesh definitions and lighting conditions.


\section{Literature Review}
In terms of problem formulation, most recent works on 3D hand modeling lean towards estimating 3D keypoints in a root-relative camera frame, centered at a hand's wrist in most cases \cite{boukhayma20193d, dkulon2019rec, Kulon_2020_CVPR, Remelli_2020_CVPR}. Very few approaches tackle the problem of getting the keypoints in the world frame \cite{MegaTrack_FB_2020}, but they require either more than one sensor or the dataset is carefully tailored that it only has image crops with the optical axis passing through the hand center \cite{spurr2020eccv}. In the existing literature, the solutions for 3D hand pose estimation vary widely depending on different factors such as types and number of sensors (RGB/grayscale, presence of lens distortion, availability of depth/infrared/IMU sensors), positioning of sensors (on the head-mounted device, wrist, hand, body or fingers), power/bandwidth/memory constraints (cloud server vs. edge device, dedicated processor vs. GPU, real-time vs. offline), the FOV coverage (single FOV vs. overlapping FOV), availability of ground-truth data (semi-supervised learning for limited 3D annotation), and even due to performance goals for user experience (high frame rate, low keypoint jitter). Another set of ideas makes use of Generative Adversarial Networks to learn more and more realistic poses and hand motions from available 3D mocap and depth sensors \cite{kocabas2019vibe}.

3D pose estimation solutions can be categorized broadly as model-free and model-based. Model-free methods for 3D pose estimation are usually an extension of 2D hand pose estimation by predicting an extra heatmap for normalized depth or distance. To provide outputs in real-world/camera space, \cite{spurr2020eccv} uses additional biomechanical constraints and an additional branch of fully connected layers to predict real depth. 
Model-based methods rely on a predefined mesh model, which could be either parametric, like MANO \cite{MANO:SIGGRAPHASIA:2017} or non-parametric, like Convolutional Mesh Autoencoder (CoMA) \cite{dkulon2019rec} and SpiralConv \cite{Kulon_2020_CVPR}. Most approaches are still confined to estimating root-relative outputs. Recently, some researchers investigated joint optimization for 2D and 3D hand pose. For example, \cite{Moon_2020_ECCV_I2L-MeshNet} demonstrated that an intermediate 2D cue (like a heatmap or lixel) can boost performance when used as a guiding mechanism to estimate 3D keypoints and meshes. However, in another work, Boukhayma et.~al \cite{boukhayma20193d} showed that using 2D heatmap guidance is good only for some specific datasets and not as much for others. For the stereo setups, hand pose estimation has been directly inspired by multi-view human pose estimation methods. Some approaches directly rely on basic triangulation as a prior followed by classic inverse kinematics to fit a 3D hand model across 2D predictions from different views \cite{MegaTrack_FB_2020}. Others rely heavily on the fusion of features in a 2D map space \cite{epipolartransformers}. A very interesting set of ideas also incline towards obtaining rotation-invariant canonical features \cite{Remelli_2020_CVPR}. Even though no present work makes use of any meta-information for 3D hand pose estimation, in \cite{Facil_2019_CVPR}, the authors used canonical camera information to improve depth estimates from monocular RGB input. In our proposed framework, instead of feeding a hand-tailored canonical representation, we pass extrinsic and intrinsic camera parameters directly as meta-information. 

\section{Hand Tracking System Setup}

In this work, our goal is to estimate the 3D keypoints and 3D hand mesh in real-world coordinates from 2D input images. We are assuming that an efficient hand detector is available which provides us both the class information and 2D bounding box for all hands in each camera frame. In our particular system setup, we have two grayscale cameras with fisheye distortion placed spatially apart but with overlapping FOVs on a head-mounted device (HMD). We assume that the cameras' intrinsic and extrinsic parameters are known. For this particular experiment, we confine ourselves to estimating only the ego-centric hand poses. Therefore, after a detector provides the bounding boxes, we can crop the region of interest (i.e. pixel representation of the 2D hands) to use as input to our system. Also, we have the bounding box information and camera parameters readily available to be used as metadata. The output of our system provides 3D keypoints and 3D Hand Mesh for the two ego-centric hands in real-world coordinates. Additionally, the system returns a compact parametric definition of the 3D hand.

\begin{figure*}[t!]
    \centering\includegraphics[width=0.7\linewidth]{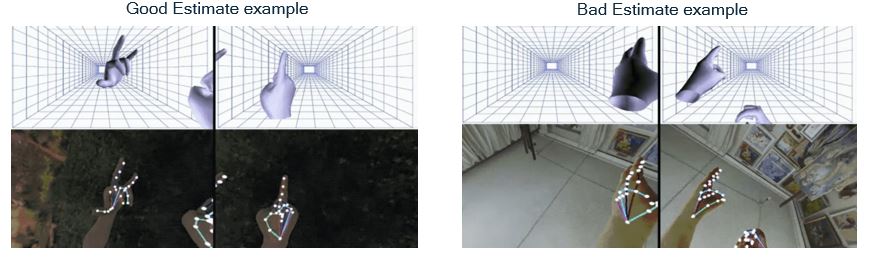}
    \caption{Good (left) and Bad (right) Estimate examples - Inference on synthetic data mimicking HMD capture.}   
    \label{fig:good_bad_cases_synth}
    \vspace{3mm}
    \centering\includegraphics[width=0.7\linewidth]{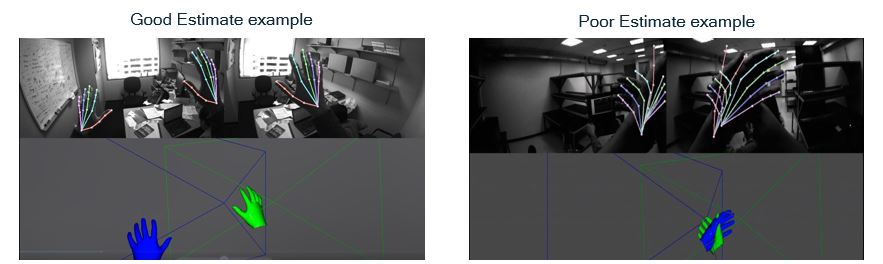}
    \caption{Good (left) and Bad (right) Estimate examples - Inference on Real Data with only 2D finetuning.}
    \label{fig:good_bad_cases_real}
    \vspace{-3mm}
\end{figure*}

\section{\textit{DIREG3D} Methodology}
Our proposed \textit{DIREG3D} framework operates in two different modes namely, monocular and stereo/multi-view. While the monocular mode is the default setup, depending on the availability of the same hand in multiple cameras, the system can easily expand to operate in the stereo or multi-view mode without any redundant computation. Detail about the modes and meta-information are further described in the following subsections.


The monocular mode of operation is illustrated in Fig.~\ref{fig:monocular_flow}. The inputs to the system in monocular mode are the $128 \times 128$ hand image crop and $28$-D normalized metadata vector. We use a CNN backbone with residual connections to extract a high-level feature from the cropped hand image, while the meta-information is passed through a separate fully connected (FC) branch. Intuitively, this step coalesces the most meaningful information from all different data modalities. After that, the fully connected layers learn to jointly optimize for multiple objectives which guide them to coherently understand 3D-2D relationships for the given lens properties. The two features are concatenated and go through another set of FC layers. These branch out to multiple heads, including MANO parameters, input for mesh convolutional decoder, and independent keypoint predictions. The mesh parameters feed into a trainable MANO model, therefore the full setup is end-to-end trainable without any pretraining required. We made a choice to use a non-parametric model as a training aid, because of the underlying flexibility it has to fit the training samples without any mathematical constraints like blending weights, and skinning or forward kinematic chains.

\subsection{Feeding Meta-Information}
Most fundamental tasks in computer vision rely heavily on the sensor information only, whereas relatively lesser exploration has been done in using some other characteristics already available as metadata. Most works that claim to infer 3D information such as depth or 3D pose from monocular input provide normalized or root-relative outputs. This is because the fundamental formulation of this problem is under-determined. We propose that some careful exploration into how these lenses distort the 3D information for 2D projection has the potential to directly give us 3D inference without training for any normalized notations. We feed the following information from our 2D detection boxes to our 3D pose network as meta-information.

\begin{itemize}
\itemsep-0.25em
    \item 1-D L2 Distance of bounding box center from optical axis. 
    \item 4-D Bounding Box corner coordinates.
    \item 1-D Scale Ratio of 2D box size to pose network input size.
    \item 9-D Rotation between the optical axis and hand box center.
    \item 9-D Modified intrinsic matrix based on the bounding box.
    \item 4-D Radial (barrel) distortion of the Fisheye Lens.
\end{itemize}

The 28-D feature vector is normalized between ranges on -1 to 1, after applying the min-max normalization based on statistics from training data for the 28 individual values. 
All of these features combined are fused with very high-level image features and the network is jointly optimized to predict the 3D keypoints and meshes. Intuitively, the meta-information enables the neural network model to predict real-world frame 3D points more accurately and helps to generalize well when trained with data from multiple camera types.

\subsection{Stereo/Multi-View Mode}
Having stereo / multi-view cameras benefits this use case in more than one way. Firstly, it provides more information about the same hand from a different but known viewpoint; secondly, it enables a wider field of view for the application as a whole. Our pipeline for leveraging stereo view to further refine the quality of the hand pose is shown in Fig.~\ref{fig:pipeline_flow}. Along with concatenating the features from two views, we also make sure that the network is aware of the extrinsic relationship between the two cameras. But, instead of passing the original camera extrinsic parameters, we first calculate the rotation and translation between the two virtual cameras, such that each virtual camera is looking at the center of detected bounding boxes from the two views. 

\subsection{Losses}
Our loss function ensures that the model is jointly optimizing for a variety of crucial objectives, including bone length, bone angles, meshes, and confidence of keypoint predictions in the 3D domain. The 3D losses comprise a bone-length loss, a bone angle loss, L1 losses on keypoints/mesh, keypoint variance estimates, and a regularization term for the MANO mesh parameters. On top of that, we also define 2D estimation losses on projected keypoints.

For cases with stereo overlap, we introduce an additional loss on 2D projections of the final 3D keypoint estimates across both the views. This imbibes knowledge about the stereo geometry within the learnable framework, and operates equivalently to inverse kinematics (IK)-based fitting methods during the training process.

\section{Results and Discussion}

Since capturing and annotating the datasets for real 3D hands is time-consuming and expensive, we created relevant synthetic data for this task. The synthetic data exactly mimicked the camera configurations of Snapdragon® XR2 HMD Reference Design \cite{Kona_xr2_device}. Multiple synthetic hand models and realistic backgrounds were iterated to create a total of $\approx150$k stereo images for the fisheye cameras, a pair which enables a full $180$ degree field of view. Table \ref{table:1} provides the Mean Keypoint Location Error (MKPE) on the in-house synthetic dataset. This is the mean of distances in mm between the ground truth 3D location and predicted 3D location of all the 21 keypoints in world space. We notice that for monocular only cases in synthetic data, we are getting ($\approx12$ mm) MKPE (on $\approx 13$k image). Whenever stereo overlap is available($\approx 4$k images), the performance improves further ($\approx11$ mm).

\begin{table}[h!]
\centering
\begin{tabular}{c  c  c} 
 \hline
 Method & Mean KPE (mm) & AUC (0-50 mm)\\ 
 \hline
 \textit{DIREG3D}-Mono & 12.37 & 0.755\\ 
 \hline
 \textit{DIREG3D}-Stereo & 11.39 & 0.774\\ 
 \hline
\end{tabular}
\caption{Results on the in-house synthetic dataset}
\label{table:1}
\vspace{-3mm}
\end{table}

The final 8-bit quantized model runs on the HMD device in under 9 ms for four crops combined. Four is the max number of crops required in a single forward pass, a case where both ego hands are visible in left and right cameras. With a lightweight hand detector to run on stereo frames together, the full solution is expected to run at 60 fps on the HMD\cite{Kona_xr2_device}.

Some qualitative examples as shown in Fig.~\ref{fig:good_bad_cases_synth} demonstrate that our approach works well in most cases with and without stereo overlap. And this is still without any prior information about the hand/temporal feedback. Also, with very limited finetuning on real data with just 2D annotations, we can see in Fig.~\ref{fig:good_bad_cases_real} that it still provides very reasonable estimates. As of now, the method still fails for some difficult cases like overlapping hands and self-occlusion with one's own hand. We believe that these can also be improved upon by using some temporal information and mesh intersection losses, which is currently a work-in-progress.

\section{Conclusion and Future Directions}
We present a simple yet efficient framework to estimate hand poses and full mesh from both single and multiple camera setups. By design, the proposed method can also flexibly accommodate lenses with different characteristics. We are currently working towards expanding the method to incorporate temporal feedback. Our future research goals also include robustly handling more complex scenarios such as extreme poses, self-occlusion, a wide variation of lighting, overlapped hands, and hands with wearables like rings, gloves, watches, and tattoos. 

{\small
\bibliographystyle{ieee}
\bibliography{egbib}

\begin{thebibliography}{10}\itemsep=-1pt

\bibitem{boukhayma20193d}
A.~Boukhayma, R.~d. Bem, and P.~H. Torr.
\newblock 3d hand shape and pose from images in the wild.
\newblock In {\em Proceedings of the IEEE Conference on Computer Vision and
  Pattern Recognition}, pages 10843--10852, 2019.

\bibitem{Kona_xr2_device}
Q.~dev network.
\newblock Qualcomm \mbox{S}napdragon® \mbox{XR2 HMD} reference design.
\newblock
  \url{https://developer.qualcomm.com/hardware/snapdragon-xr2-hmd-reference-design}.

\bibitem{Facil_2019_CVPR}
J.~M. Facil, B.~Ummenhofer, H.~Zhou, L.~Montesano, T.~Brox, and J.~Civera.
\newblock {CAM-Convs: Camera-Aware Multi-Scale Convolutions for Single-View
  Depth}.
\newblock In {\em The IEEE Conference on Computer Vision and Pattern
  Recognition (CVPR)}, June 2019.

\bibitem{MegaTrack_FB_2020}
S.~Han, B.~Liu, R.~Cabezas, C.~D. Twigg, P.~Zhang, J.~Petkau, T.-H. Yu, C.-J.
  Tai, M.~Akbay, Z.~Wang, A.~Nitzan, G.~Dong, Y.~Ye, L.~Tao, C.~Wan, and
  R.~Wang.
\newblock Megatrack: Monochrome egocentric articulated hand-tracking for
  virtual reality.
\newblock {\em ACM Trans. Graph.}, 39(4), July 2020.

\bibitem{epipolartransformers}
Y.~He, R.~Yan, K.~Fragkiadaki, and S.-I. Yu.
\newblock Epipolar transformers.
\newblock In {\em Proceedings of the IEEE/CVF Conference on Computer Vision and
  Pattern Recognition}, pages 7779--7788, 2020.

\bibitem{kocabas2019vibe}
M.~Kocabas, N.~Athanasiou, and M.~J. Black.
\newblock Vibe: Video inference for human body pose and shape estimation.
\newblock In {\em The IEEE Conference on Computer Vision and Pattern
  Recognition (CVPR)}, June 2020.

\bibitem{Kulon_2020_CVPR}
D.~Kulon, R.~A. Guler, I.~Kokkinos, M.~M. Bronstein, and S.~Zafeiriou.
\newblock Weakly-supervised mesh-convolutional hand reconstruction in the wild.
\newblock In {\em Proceedings of the IEEE/CVF Conference on Computer Vision and
  Pattern Recognition (CVPR)}, June 2020.

\bibitem{dkulon2019rec}
D.~Kulon, H.~Wang, R.~A. G{\"{u}}ler, M.~M. Bronstein, and S.~Zafeiriou.
\newblock Single image 3d hand reconstruction with mesh convolutions.
\newblock In {\em Proceedings of the British Machine Vision Conference
  ({BMVC})}, 2019.

\bibitem{Masurovsky2020ControllerFreeHT}
A.~Masurovsky, P.~Chojecki, D.~Runde, M.~Lafci, D.~Przewozny, and M.~Gaebler.
\newblock Controller-free hand tracking for grab-and-place tasks in immersive
  virtual reality: Design elements and their empirical study.
\newblock {\em Multimodal Technologies and Interaction}, 4(4), 2020.

\bibitem{Moon_2020_ECCV_I2L-MeshNet}
G.~Moon and K.~M. Lee.
\newblock I2l-meshnet: Image-to-lixel prediction network for accurate 3d human
  pose and mesh estimation from a single rgb image.
\newblock In {\em European Conference on Computer Vision (ECCV)}, 2020.

\bibitem{Remelli_2020_CVPR}
E.~Remelli, S.~Han, S.~Honari, P.~Fua, and R.~Wang.
\newblock Lightweight multi-view 3d pose estimation through camera-disentangled
  representation.
\newblock In {\em IEEE/CVF Conference on Computer Vision and Pattern
  Recognition (CVPR)}, June 2020.

\bibitem{MANO:SIGGRAPHASIA:2017}
J.~Romero, D.~Tzionas, and M.~J. Black.
\newblock Embodied hands: Modeling and capturing hands and bodies together.
\newblock {\em ACM Transactions on Graphics, (Proc. SIGGRAPH Asia)}, 36(6),
  Nov. 2017.

\bibitem{spurr2020eccv}
A.~Spurr, U.~Iqbal, P.~Molchanov, O.~Hilliges, and J.~Kautz.
\newblock Weakly supervised 3d hand pose estimation via biomechanical
  constraints.
\newblock In {\em European Conference on Computer Vision (ECCV)}, 2020.

\bibitem{issc18}
K.~L. Sungpill~Choi, Jinsu~Lee and H.-J. Yoo.
\newblock A 9.02mw cnn-stereo-based real-time 3d hand-gesture recognition
  processor for smart mobile devices.
\newblock {\em International Solid-State Circuits Conference (ISSCC)}, 2018.

\bibitem{taylor2016efficient}
J.~Taylor et~al.
\newblock Efficient and precise interactive hand tracking through joint,
  continuous optimization of pose and correspondences.
\newblock {\em ACM Transactions on Graphics (TOG) - Proceedings of ACM SIGGRAPH
  2016}, 35, July 2016.

\end{thebibliography}
}

\end{document}